\documentclass{article}

\usepackage{arxiv}

\usepackage{lineno} 

\usepackage{times}
\usepackage{latexsym}

\usepackage[T1]{fontenc}

\usepackage[utf8]{inputenc}

\usepackage{microtype}

\usepackage{inconsolata}

\usepackage{graphicx} 
\usepackage{dsfont}
\usepackage{enumitem}
\usepackage{pgfplots}
\usepackage{subcaption}
\usepackage{booktabs}
\usepackage{float}
\usepackage{xspace}
\usepackage{multirow}

\usepackage{balance}
\pgfplotsset{compat=newest}
\usepackage[listings,breakable,skins]{tcolorbox}
\usepackage{refcount}
\usepackage{stfloats} 
\usepackage{tablefootnote}
\usepackage{url}
\usepackage{nameref}

\newcommand{\Comment}[1]{}

\newcommand{\Ni}{({\em i})~}
\newcommand{\Nii}{({\em ii})~}
\newcommand{\Niii}{({\em iii})~}
\newcommand{\Niv}{({\em iv})~}

\newcommand{\liverag}{{\em LiveRAG}}
\newcommand{\acs}{{\em ACS}}
\newcommand{\diff}{{\em diff}}
\newcommand{\disc}{{\em disc}}

\begin{document}

\title{\liverag{}: A diverse Q\&A dataset with varying difficulty level for RAG evaluation}
\author{
David Carmel, Simone Filice, Guy Horowitz, Yoelle Maarek, Alex Shtoff, Oren Somekh, Ran Tavory\\
Technology Innovation Institute (TII), Haifa, Israel
}


\maketitle

\begin{abstract}
 With Retrieval-Augmented Generation (RAG) becoming more and more prominent in generative AI solutions, there is an emerging need for systematically evaluating their effectiveness.
 We introduce the \liverag{} benchmark, a publicly available dataset of 895 synthetic questions and answers designed to support systematic evaluation of RAG-based Q\&A systems. This synthetic benchmark is derived from the one used during the SIGIR'2025 \liverag{} Challenge, where competitors were evaluated under strict time constraints. 
 It is augmented with information that was not made available to competitors during the Challenge, such as the ground-truth answers, together with their associated supporting claims which were used for evaluating competitors' answers. 
 In addition, each question is associated with 
 estimated difficulty and discriminability scores, derived from applying an Item Response Theory model to competitors' responses.
 Our analysis highlights the benchmark’s questions diversity, the wide range of their difficulty levels, and their usefulness in differentiating between system capabilities. The \liverag{} benchmark will hopefully help the community advance RAG research, conduct systematic evaluation, and develop more robust Q\&A systems.
\end{abstract}

\section{Introduction}

\textit{Retrieval-Augmented Generation} (RAG) is a widely adopted methodology for improving the effectiveness of \textit{Large Language Models} (LLMs), particularly for question answering tasks \cite{lewis2020retrieval,izacard2022few, Guo+al:23a}. RAG is attracting significant attention from the AI and Information Retrieval (IR) communities. Yet, reliable and systematic evaluation of RAG systems remains an open challenge \cite{es2024ragas, yang2024crag, thakur2025support}. 

In this paper, we introduce a publicly available benchmark for evaluating RAG-based question-answering systems. The ``\liverag{} benchmark'' we release in this work\footnote{
\url{https://huggingface.co/datasets/LiveRAG/Benchmark}
}
is derived from the one used during the SIGIR-2025 \liverag{} Challenge \cite{carmel2025sigir2025liverag}, hence its name. 

The SIGIR \liverag{} Challenge took place between March and May 2025, with results announced during the SIGIR'2025 conference. Its goal was to facilitate progress in RAG research by enabling teams from academia and industry to evaluate their solutions on a common benchmark  and compare performance against those of other teams, using a fixed external corpus (Fineweb-10BT\footnote{\url{https://huggingface.co/datasets/HuggingFaceFW/fineweb/viewer/sample-10BT}}), and a fixed open-source LLM (Falcon3-10B-Instruct\footnote{\url{https://huggingface.co/tiiuae/Falcon3-10B-Instruct}}).
During the live day event, competing teams were divided into two sessions, each receiving a set of 500 unseen questions, including 105 questions shared between sessions, for manual validation of LLM-based judgment and cross-session calibration. 
All questions (and associated reference answers) were generated using the DataMorgana tool \cite{filice2025dmacl} (see \S{\ref{sec:benchmark}} for more details).



To generate the \liverag{} benchmark we introduce here, we merged the two sessions' sets of
500 questions (with their shared 105 questions) to obtain a total of 895 unique questions. We then augmented these questions with supplementary information that was not made available to competitors, thereby enabling multiple and richer evaluation scenarios.

The \liverag{} benchmark provides, for each question, the answer generated by DataMorgana, as well as the supporting documents that the tool used for Q\&A generation. It also includes the ``answer claims'' used during the challenge to compare competitors' answers with the reference answers. Furthermore, it associates with each question an estimated \textit{difficulty score}, and a \textit{discriminability score} derived from an Item Response Theory (IRT) model \cite{lord2012applications} trained on the evaluation scores of participating systems' responses to the \liverag{} questions\footnote{
We thank the organizers of the SIGIR'25 LiveRAG Challenge for giving us access to the participant answer scores for each question, which enabled us to compute the IRT model parameters.
}. 



The IRT-derived difficulty and discriminability parameters provided by the benchmark serve to normalize question characteristics across the dataset, effectively placing them on a common scale. This calibration is essential since the questions are distributed across two disjoint sessions, with responses originating from distinct participant cohorts.
Furthermore, these parameters enable practitioners to train their systems using questions of varying difficulty levels, e.g., for curriculum learning \cite{soviany2022curriculum}.
Our analysis demonstrates that these parameters effectively reflect question difficulty and discriminability, as questions with lower difficulty values were consistently more challenging for all RAG-based systems that participated in the challenge, as well as for a wide range of LLMs of varying sizes.
  
The remainder of this paper is organized as follows. Section~\S\ref{sec:benchmark} describes the process used to construct the benchmark. 
In Section \ref{sec:irt-model} we describe the IRT model used for benchmark analysis.
Section~\S\ref{sec:analysis} presents an analysis of the questions’ difficulty, followed by Section~\S\ref{sec:reasons}, which explores factors contributing to question difficulty. Finally, Section~\S\ref{sec:limitations} discusses limitations of the benchmark, and Section~\S\ref{sec:conclusions} concludes.

\section{Benchmark Generation with DataMorgana}
\label{sec:benchmark}
The \liverag{} Benchmark was generated using DataMorgana~\cite{filice2025dmacl}, a synthetic data generation tool that offers high configurability and is capable of producing highly diverse sets of 
Q\&As. The following section highlights some of DataMorgana's characteristics that were specifically leveraged to produce the \liverag{} benchmark.

\subsection{Document sampling}\label{sec:dm_input}


DataMorgana generates each QA pair based on information extracted from specific source documents. To construct the \liverag{} benchmark, documents were sampled from the official corpus of the Challenge, FineWeb-10BT. 
Given that the corpus comprises arbitrary web pages, a topic-based document sampling pipeline was employed to ensure the selected documents are appropriate for generating valuable question-answer pairs.
The sampling pipeline comprises three stages:  

{\bf Topic Generation.} The LLM is prompted to generate diverse list of high-level topics, and for each topic, a list of related subtopics (See Appendix \S{\ref{appendix:prompts:topic_generation}} for the topic generation prompt). 

{\bf Topic-Based Document Retrieval.} The subtopics are used to query FineWeb-10T for retrieving relevant documents. 

{\bf Document Filtering.} Duplicate, too short, or too long documents are removed from the pool of retrieved documents. 
We then use an LLM to score each document according to the following criteria:
\begin{itemize}[leftmargin=0.2cm]
\item \emph{Factuality} — Does the document contain factual information that is appropriate for generating open-domain questions? 
\item \emph{Interest} — Is the content potentially interesting and useful? 
\item \emph{Credibility} — Is the document trustworthy and free from promotional narrative, or overly subjective material? 
\item \emph{Toxicity} — Does the document contain harmful, offensive, or inappropriate language? 
\item \emph{Sexuality} — Does the document contain sexual content? 
\item \emph{Freshness}  — Is the content fresh and relevant, or is it outdated? 
\end{itemize} 
The documents are filtered according to their scores using predefined thresholds for each criterion, 
constructing a pool of valid documents to be used for question generation. The filtering prompt is given in Appendix \S{\ref{appendix:prompts:document_filtering}}.

\subsection{Generation pipeline} 
\label{sec:dm_pipeline}
DataMorgana builds the benchmark incrementally, generating one Q\&A pair at a time, using the following three-step procedure.

\subsubsection{Category set selection} 
\label{sec:dm_config}
The benchmark is defined by specifying a set of desired question categorizations, each comprising one or more mutually exclusive categories. For \liverag{}, eight such categorizations were used, as listed in Table \ref{tab:question categories}. These categories are intentionally broad to support the generation of diverse questions from any document within the corpus. In each question generation task, one category is randomly selected from each categorization, resulting in a total of eight categories per task.    
The 8 selected categories are used for the question generation step (see \S{\ref{par:llm_invocation}}).
    
    \subsubsection{Document selection} 
    DataMorgana randomly samples  a document from the pool of valid documents, to be used for Q\&A generation. A second, complementary document is selected for multi-doc generation, if either {\em comparison} or {\em multi-aspect} has been selected from the {\em Answer Type} categorization. The second document is selected by prompting the LLM to generate 3 questions that can be partially answered by the first selected document, $d$, and for each question, generating a search query for retrieving the missing information not presented in $d$ (See Appendix \S{\ref{appendix:prompts:query_generation}} for the relevant prompt).
    Then, five documents are retrieved from the corpus for each generated query, and the LLM is prompted to select one  document from the pool of search results that best complements $d$. The prompt for selecting the complementary document is given in Appendix \S{\ref{appendix:prompts:document_selection}}.
    
    \subsubsection{Question generation}
    \label{par:llm_invocation} 
    Finally, a prompt is constructed to instruct the LLM to generate a Q\&A from the selected document (and from the complementary document in the case of {\em comparison} or {\em multi-aspect} questions), ensuring that the generated Q\&A adheres to the eight selected categories (See the question generation prompt in \cite[Appendix A]{filice2025dmacl}).
For all generation tasks, we used Claude 3.5-Sonnet\footnote{\url{https://www.anthropic.com/news/claude-3-5-sonnet}} as the backbone LLM.

The LiveRAG benchmark comprises 895 question-answer pairs generated through the process outlined above. A comprehensive description of the dataset, including a few Q\&A examples,  is provided in Appendix \S{\ref{app:description}}.


\begin{table*}[ht!]
\scriptsize
\centering
\begin{tabular}{c|c|p{11cm}} 
\hline
\textbf{Categorization} & \textbf{Category} & \textbf{Description} \\ \hline
\multirow{17}{*}{Answer Type} & \multirow{2}{*}{factoid} & question seeking a specific, concise piece of information or a short fact about a particular subject, such as a name, date, or number. \\ \cline{2-3}
 & \multirow{1}{*}{yes/no} & a question that can be answered with true/false or yes/no.\\ \cline{2-3}
 & \multirow{1}{*}{definition} & a question that requires finding the definition of the term in the question.\\ \cline{2-3}
 & \multirow{1}{*}{list} & a question that requires as an answer a list of entities or facts.\\ \cline{2-3}
 & \multirow{1}{*}{explanation} &  a question that requires as an answer an explanation.\\ \cline{2-3}
 & \multirow{6}{*}{comparison} & a comparison question that requires comparing two related concepts or entities. The comparison must be natural and reasonable, i.e., comparing two entities by a common attribute which is meaningful and relevant to both entities. For example: 'Who is older, Glenn Hughes or Ross Lynch?', 'Are Pizhou and Jiujiang in the same province?', 'Pyotr Ilyich Tchaikovsky and Giuseppe Verdi have this profession in common'. The information required to answer the question needs to come from two documents, specifically, the first document must provide information about the first entity/concept, while the second must provide information about the second entity/concept.\\ \cline{2-3}
  & \multirow{5}{*}{multi-aspect} & a question about two different aspects of the same entity/concept. For example: 'What are the advantages of AI-powered diagnostics, and what are the associated risks of bias in medical decision-making?', 'How do cryptocurrencies enable financial inclusion, and what are the security risks associated with them?'. The information required to answer the question needs to come from two documents, specifically, the first document must provide information about the first aspect, while the second must provide information about the second aspect.\\ \hline

\multirow{5}{*}{Answer Style} & \multirow{2}{*}{concise-answer} & a question that explicitly asks for a short and direct answer, requesting only the essential information without additional explanation. The question must include this instruction explicitly. \\ \cline{2-3}
 & \multirow{2}{*}{detailed-answer} & a question that explicitly asks for a comprehensive answer, requesting additional details, background information, or clarifications. The question must include this instruction explicitly.\\ \cline{2-3}
  & \multirow{1}{*}{unspecified} & a question that does not explicitly ask for a specific style of answer.\\ \hline

\multirow{2}{*}{Premise} & \multirow{1}{*}{direct} & question that does not contain any premise or any information about the user \\ \cline{2-3}
& \multirow{1}{*}{with-premise} & question starting with a very short premise, where the user reveals their needs or some information about himself. \\ \hline

\multirow{6}{*}{Phrasing} & \multirow{1}{*}{concise-and-natural} & a concise direct natural question consisting of a few words.\\ \cline{2-3}
& \multirow{1}{*}{verbose-and-natural} & a relatively long question consisting of more than 9 words.\\ \cline{2-3}
& \multirow{2}{*}{short-search-query} & phrased as a typed web query for search engines (only keywords, without punctuation and without a natural-sounding structure). It consists of less than 7 words.\\ \cline{2-3}
& \multirow{2}{*}{long-search-query} & phrased as a typed web query for search engines (only keywords, without punctuation and without a natural-sounding structure). It consists of more than 6 words. \\ \hline

\multirow{3}{*}{Linguistic Variation}  & \multirow{1}{*}{similar-to-document} & a question phrased using the same terminology and phrases appearing in the document. \\ \cline{2-3}
 & \multirow{2}{*}{distant-from-document} & a question phrased using expressions and terminology that differ from the ones appearing in the documents. For instance, the question uses complex paraphrasing that deviates from the original document's exact wording.\\ \hline  

\multirow{2}{*}{Politeness}  & \multirow{1}{*}{polite} & a question phrased in a very polite way. \\ \cline{2-3}
 & \multirow{1}{*}{neutral} & a question that is not rude but that at the same time does not include social niceties like 'please' or 'would you mind'.\\ \hline  

\multirow{3}{*}{Linguistic Correctness}  & \multirow{1}{*}{correct} & a question written in correct English. \\ \cline{2-3}
& \multirow{1}{*}{mild-mistakes} & a question containing mild spelling mistakes.\\ \cline{2-3}
& \multirow{1}{*}{severe-mistakes} & a question containing severe spelling mistakes.\\ \hline  

\multirow{4}{*}{User Persona}  & \multirow{1}{*}{expert} & an expert on the subject discussed in the documents, therefore, he asks complex questions. \\ \cline{2-3}
& \multirow{1}{*}{novice} & a person with very basic knowledge on the topic discussed in the topic. Therefore, he asks very simple questions.\\ \cline{2-3}
& \multirow{1}{*}{researcher} & a researcher operating on the topic discussed in the documents.\\ \cline{2-3} 
& \multirow{1}{*}{journalist} & a journalist interested in writing an article about the topic discussed in the documents.\\ \hline  
 
\end{tabular}
\caption{DataMorgana configuration for Question Categorizations and for User Personas, used for generating the LiverRAG benchmark.}  
\label{tab:question categories}
\end{table*}

\section{IRT analysis}
\label{sec:irt-model}

\subsection{Background}
We analyze the benchmark characteristics using \textit{Item Response Theory} (IRT), a framework from psychometrics \cite{lalor2018understanding, rodriguez2021evaluation, vania021comparing},  
which jointly estimates latent traits of questions and of participating systems (subjects) in the \liverag{} challenge. 

IRT is frequently used in educational testing \cite{lord2012applications}, as well as in machine learning \cite{smith2014instance, lorena2024trusting}. Recent work investigates its use for dataset analysis \cite{vania021comparing, rodriguez2021evaluation}. We follow this line of work to analyze the \liverag{} benchmark, and expose IRT model parameters as part of the dataset, thus enabling practitioners to train their systems using questions of varying difficulty levels. 

Given an observation matrix $Y^{m\times n}$ of $m$ subjects and $n$ questions, where $Y[j,i]$ represents the correctness score of the answer provided by subject $s_j$ to question $q_i$\footnote{The observation matrix is not necessarily complete, i.e., a subject may answer only a subset of the questions.}; an IRT model estimates the probability of $s_j$ answering $q_i$ correctly by learning the latent parameters of $s_j$ and $q_i$ that best fit the input observation data.

A series of statistical models with increasing complexity are used to represent both item and subject characteristics. The IRT one-parameter logistic model (1PL), also known as the Rasch model, estimates a latent ``skill'' parameter $\theta_j$ for each subject, and a latent ``difficulty'' parameter $b_i$ for each question. It is defined by:
\begin{equation}
\label{eq:1pl} 
p(y_{j,i} = 1|\theta_j,b_i)= \frac{1}{1 + e^{-(\theta_j-b_i)}}
\end{equation}
The larger the margin between $\theta_j$ and $b_i$ is, the higher the probability of $s_j$ answering $q_i$ correctly. 

More complex IRT models estimate additional latent parameters for items. The two-parameter logistic (2PL) model introduces a ``discrimination'' parameter $a_i$ for each question $q_i$, which reflects how effectively the question discriminates between individuals with similar skills: 
\begin{equation}
\label{eq:2pl}
p(y_{j,i} = 1|\theta_j,b_i, a_i)= \frac{1}{1 + e^{-a_i(\theta_j-b_i)}}
\end{equation}
Other IRT models that include a ``guessing'' parameter for each question, or multi-dimensional parameters 
\cite{Lalor2023-py-irt}, are outside the scope of this work.

\subsection{IRT model implementation}
To implement the IRT models, we use the py-irt package\footnote{\url{https://github.com/nd-ball/py-irt}}~\cite{Lalor2023-py-irt}, a Python package based on probabilistic inference for fitting the latent subject and item parameters that best explain the observed data. 
For observations, we leverage the Correctness metric~\cite{carmel2025sigir2025liverag} used for evaluating the system's answer for a given question in the \liverag{} Challenge.  The Correctness score is defined as the harmonic mean of {\em Coverage} and {\em Relatedness}, with Coverage being the proportion of critical content in the reference answer that is correctly reflected in the generated answer, and Relatedness being the proportion of vital claims in the generated answer that are relevant to the given question\footnote{
The py-irt package expects binary observations (true or false), whereas in our case, observations are continuous in the range of $[-1\ldots,2]$, modeling the extent to which the answer is correct. We therefore modified the package to support continuous observations by using the Continuous-Bernoulli distribution for the observation likelihood, rather than the Bernoulli distribution originally used by the package. Since this distribution expects observations in the range $[0\ldots1]$, we linearly transformed the Correctness scores to this range.

}.

\begin{figure}[t]
    \centering
    \includegraphics[width=0.48\textwidth]{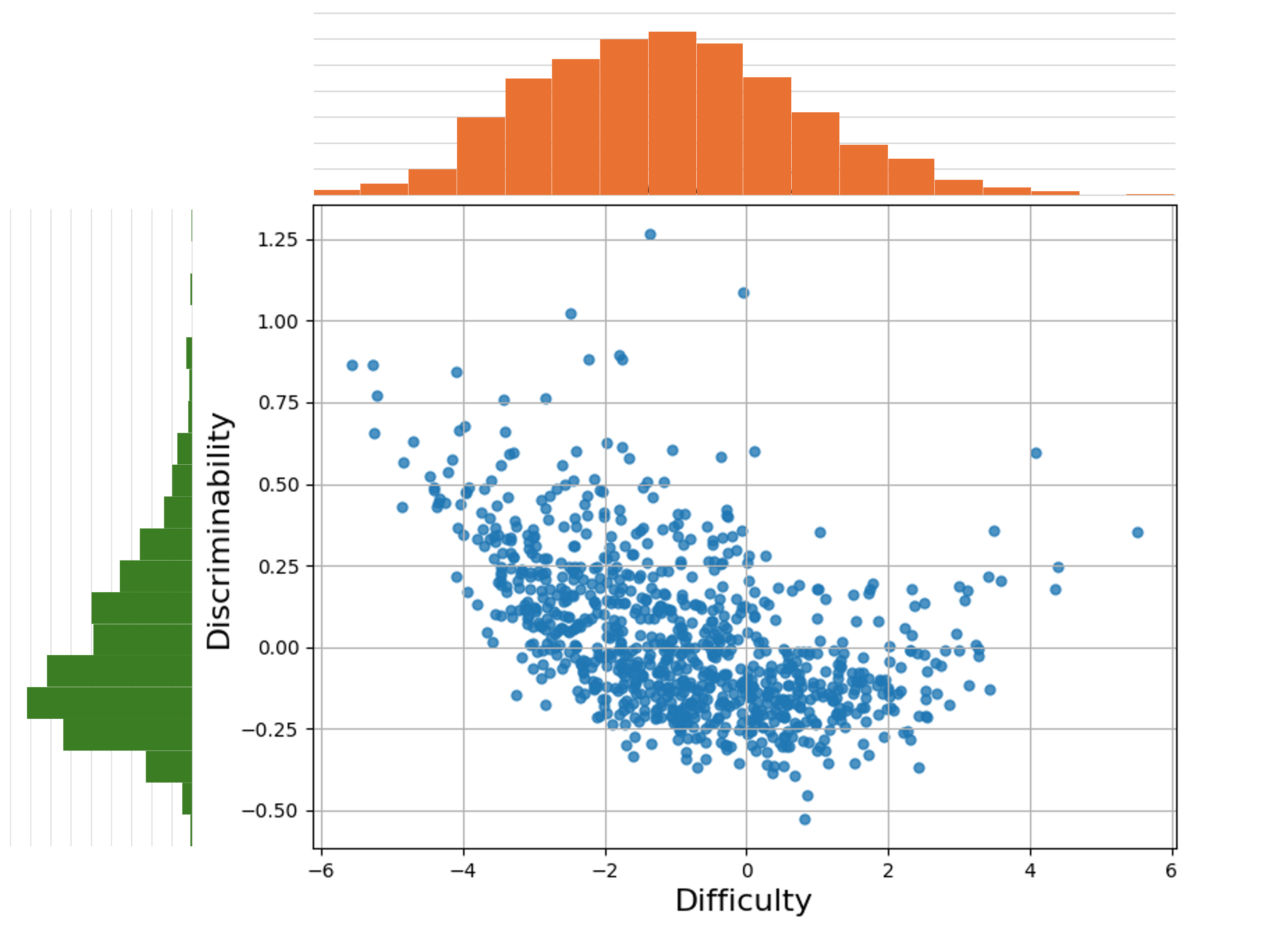}
     \caption{Question parameters learned by the IRT-2PL model for the \liverag{} dataset. 
     Top: Difficulty distribution. Left: Discriminability distribution. Middle: Scatter plot of difficulty and discriminability scores of all benchmark questions. 
     }
     \label{fig:2pl}
\end{figure}

In this work we focus on the 2PL model.
Training was conducted with a learning rate of 0.01, dropout=0.2 and over 10,000 epochs.
The parameters learned by the model, $(b_i, a_i)$, per question $q_i$, are provided as part of the benchmark. Figure \ref{fig:2pl} presents the question difficulty (\diff{}) and discriminability (\disc{}) distributions, alongside a scatter plot of $(b_i,a_i)$ of all questions. 
Interestingly, the Pearson correlation between \diff{} and the average correctness scores (\acs{}) (the average Correctness score of all participating systems that answer the question) is -0.97\footnote{Correlation between \diff{} and \acs{} is negative since high Correctness score indicates low difficulty.}. 
Overall, there is a weak negative correlation between discrimination and difficulty (Pearson = -0.423).

Furthermore, when comparing system rankings, derived from the learned skills ($\theta_j$), with their leaderboard positions reported in \cite{carmel2025sigir2025liverag}, we observe a strong concordance -- reflected by Kendall’s tau coefficients of 0.766 for the first session and 0.999 for the second. This high correlation persists despite the fact that skill-based rankings are computed over the full benchmark set, whereas leaderboard scores are session-specific, each based on a subset of 500 questions.


\section{Validating Question Difficulty}
\label{sec:analysis}
\begin{figure}[t]
    \centering
     \includegraphics[width=0.48\textwidth]{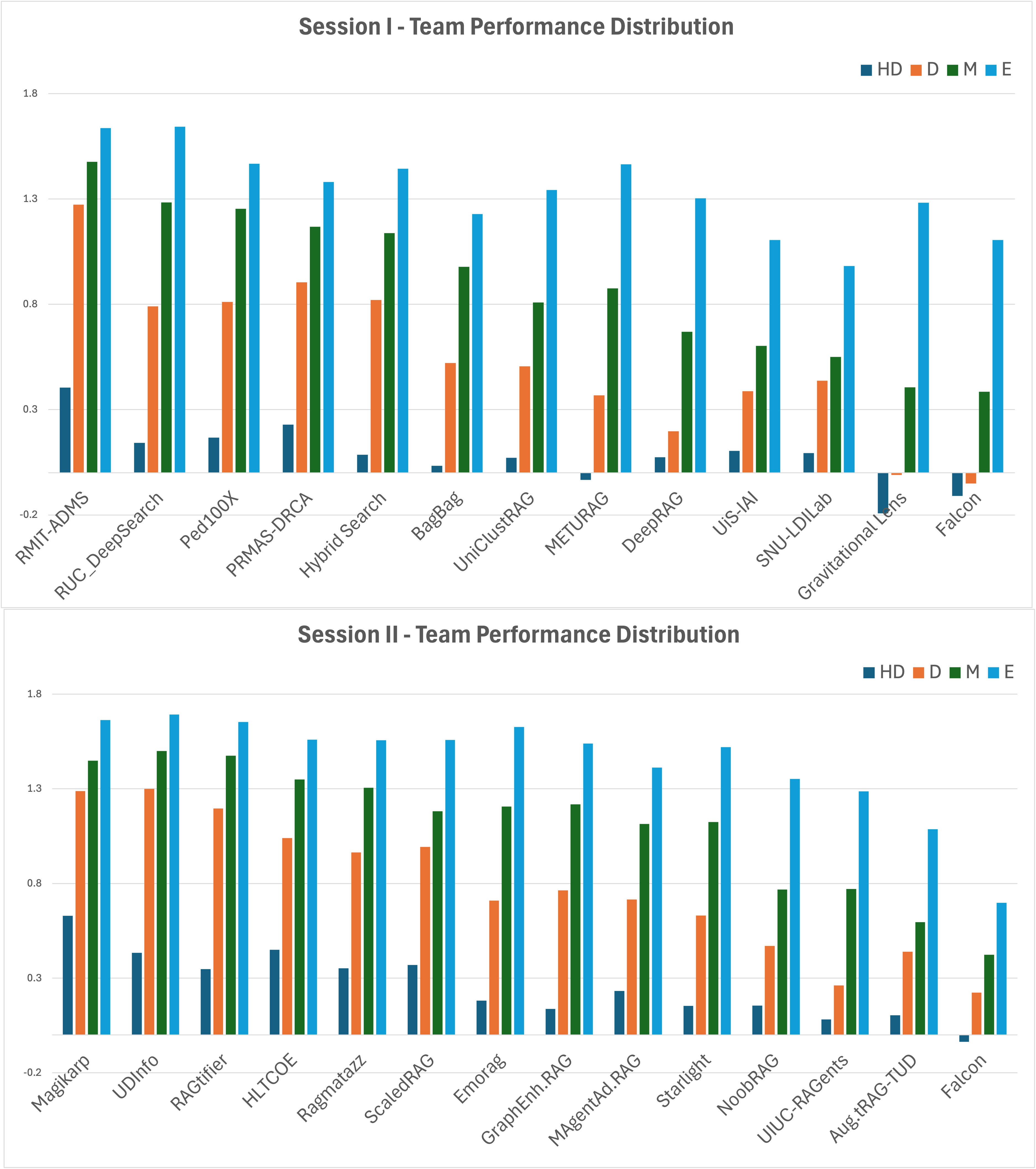}
     \caption{\label{fig:diffDistr} Team performance distributions across \diff{} bins. Teams are ordered from left to right by their leaderboard position. The rightmost distribution represents Falcon3-10B without RAG, given for reference.}
\end{figure}

To validate the effectiveness of the difficulty scores learned by the IRT model in estimating question difficulty, we analyze the distribution of \diff{} scores across the \liverag{} questions. For clarity, we divide the questions into quartiles according to their \diff{} score:
\Ni HD (highly difficult) in the range $[-6, -2.143)$; \Nii D (difficult) in the range $[-2.143, -0.962)$; \Niii M (moderate) in the range $[-0.962, 0.236)$; and \Niv E (easy) $[0.236, 6]$.
Figure \ref{fig:diffDistr} illustrates the 
performance distribution of all systems participating in the \liverag{} challenge over the \diff{} bins. We order the systems, from left to right, according to their official leaderboard score.

Examining the graphs, we see that for both sessions and for all systems, performance consistently improves as questions become easier. This trend confirms that the \diff{} scores accurately reflect question difficulty. It is also observed that in both sessions, Falcon3 without RAG underperforms as compared to all participating systems that used a RAG-based solution. This highlights the long-tail nature of the benchmark questions, which often require retrieval assistance to be answered effectively.

These findings are further substantiated by Table \ref{tab:QDiss}: ACS exhibits a monotonic increase from harder to easier bins, aligning with expectations. The average {\it disc} score declines across bins, indicating that hard questions provide weaker discriminatory power to the benchmark. 

\begin{table}[ht]
    \centering
    \addtolength{\tabcolsep}{-0.1em}
    \begin{tabular}{ccccc}
                     &HD     &D      &M      &E \\
    \toprule
    \#questions     &224    &223    & 224      & 224 \\
    {\it ACS}      &-1.304  &-0.653 &-1.883  &2.325 \\
    {\it disc}     &0.230   &0.051  &-0.041    &-0.106\\
    \bottomrule
    \end{tabular}
    \caption{{\it ACS} and {\it disc} scores across \diff{}  bins.}
    \label{tab:QDiss}
\end{table}

\subsection{Are difficult questions difficult for all?}

\begin{figure}[t]
    \centering
     \includegraphics[width=0.48\textwidth]
     {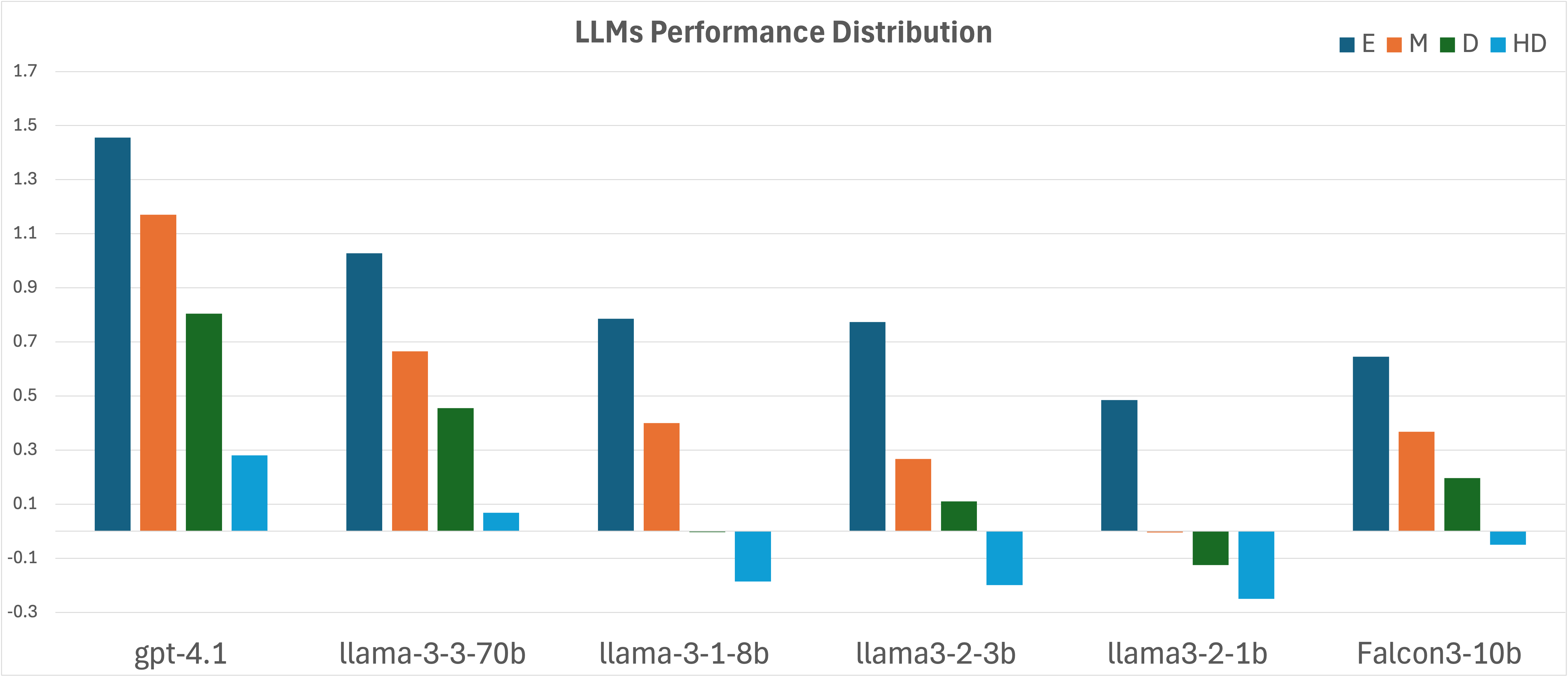}
     \caption{\label{fig:gptllama}
        Performance distributions of GPT-4.1, and several LLaMA models of different sizes (without RAG), across the {\em diff} bins.
     }
\end{figure}

We evaluate GPT-4.1\footnote{GPT-4.1 version: gpt-4.1-2025-04-14} answers over the Benchmark questions, alongside several LLaMA models of varying sizes\footnote{
\url{https://huggingface.co/meta-llama/Llama-3.3-70B-Instruct},
\url{https://huggingface.co/meta-llama/Llama-3.1-8B-Instruct},
\url{https://huggingface.co/meta-llama/Llama-3.2-3B-Instruct},
\url{https://huggingface.co/meta-llama/Llama-3.2-1B-Instruct}
},
to examine whether difficult questions are consistently challenging across models. 
We applied the same LLM-as-a-judge \cite{gu2024survey} process used in the LiveRAG challenge to measure ACS of the LLM responses to the challenge questions.
Figure~\ref{fig:gptllama} presents the average performance of the LLMs (without RAG augmentation) across the predefined {\it diff} bins. The results reveal that question difficulty strongly correlates with model performance, i.e., harder questions yield lower average ACS score irrespective of model architecture or size. Furthermore, the performance order between difficulty levels remains consistent across the evaluated LLMs. As expected, larger models outperform smaller ones. 
Interestingly, GPT-4.1 surpasses some participating systems, yet it is outperformed by the top-performing \liverag{} teams that implemented RAG on top of Falcon3-10B. This supports our observation that in the absence of RAG, even state-of-the-art LLMs struggle to answer the benchmark’s questions effectively, underscoring RAG necessity for long-tail questions.

\section{Analyzing Question Difficulty}
\label{sec:reasons}
What factors make a question difficult for LLMs, and more specifically, for RAG-based LLMs? Several factors can increase difficulty, including surface-level issues such as severe typographical errors or lack of clarity, as well as deeper challenges requiring complex reasoning. Moreover, question-independent factors such as inaccuracies or omissions in the reference answer, an absence of relevant content in the RAG corpus, cascading errors from retrieval, or limited coverage of certain domains by the LLM, can also impact perceived difficulty \cite{sugawara2022makes, liu2022challenges}. In this section, we focus exclusively on question-intrinsic difficulty, leaving corpus- and system-level effects to future work. To this end, we analyze  the IRT-derived \diff{} scores across various question types to better understand the structural and semantic properties that drive question difficulty in RAG-based systems.

\subsection{Single- vs. Multi-Document questions}
Single-doc questions are generated by DataMorgana using a single source document, and it is guaranteed that each question can be answered by that document\footnote{We note that while alternative, better answers based on other documents in the corpus may exist, the selected document is guaranteed to answer the question.}. In contrast, multi-doc questions (i.e., \texttt{comparison} or \texttt{multi-aspect} questions) are generated from two complementary documents (see Section \ref{sec:dm_pipeline}). As such, many of these questions cannot be accurately answered using a single document alone.

Therefore, we hypothesize that multi-doc questions are more difficult than single-doc questions. Table \ref{tab:singleVSmulti} supports this hypothesis by reporting the average {\it diff} and {\it disc} scores for the sets of single and multi-doc questions. As expected, multi-doc questions exhibit a significantly higher {\it diff} score than single-doc questions,  while showing lower discriminative power as reflected by lower average {\it disc} score.
\begin{table}[ht]
    \centering
    \begin{tabular}{cccc}
            &\#questions     & {\it diff}       &{\it disc}\\
       \toprule
    Single-doc  &758             &-1.083   &0.062\\
    Multi-doc   &137             &0.091   &-0.212\\
    \bottomrule
    \end{tabular}
    \caption{Average {\it diff} and {\it disc} scores across single- and multi-doc questions.}
    \label{tab:singleVSmulti}
\end{table}

\subsection{Difficulty across Question Categories}

The DataMorgana configuration categories used for question generation can also be related to question difficulty. For instance, questions containing severe linguistic typos are likely to pose greater challenges for a question answering system.

\begin{figure*}[ht]
    \centering
    \begin{subfigure}[b]{0.33\textwidth}
        \includegraphics[width=\textwidth]{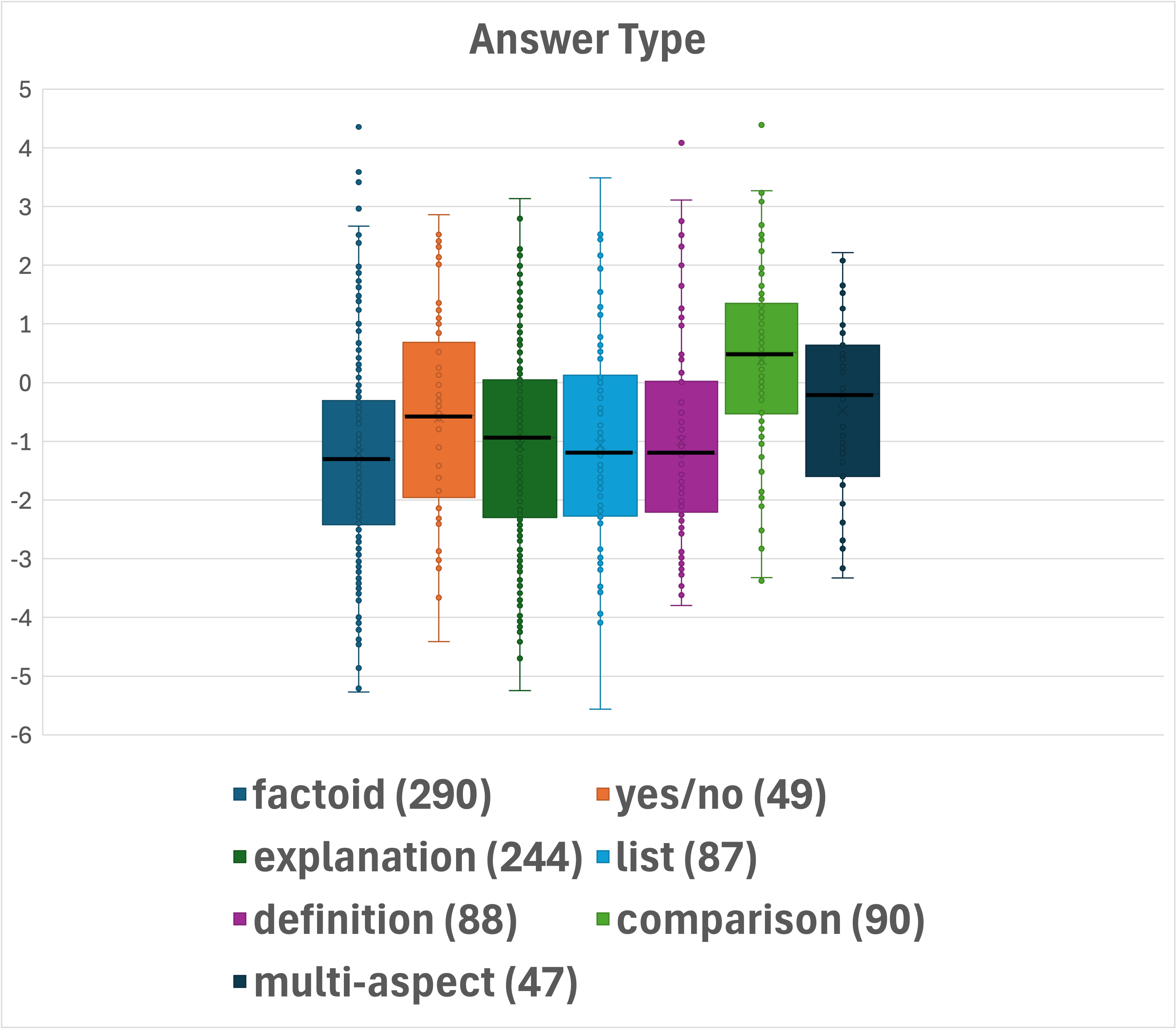}
    \end{subfigure}%
    \begin{subfigure}[b]{0.33\textwidth}
        \includegraphics[width=\textwidth]{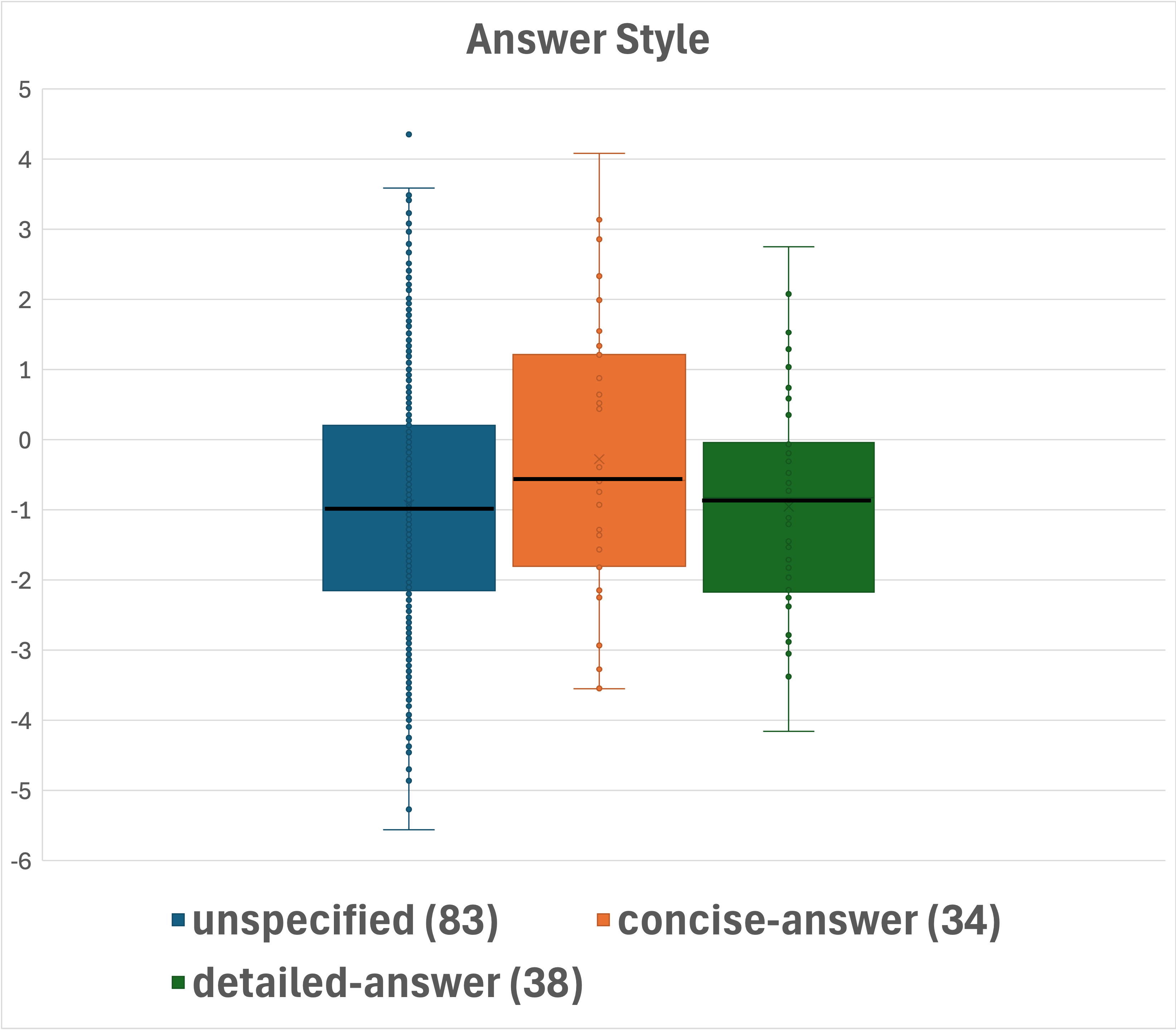}
    \end{subfigure}%
    \begin{subfigure}[b]{0.33\textwidth}
        \includegraphics[width=\textwidth]{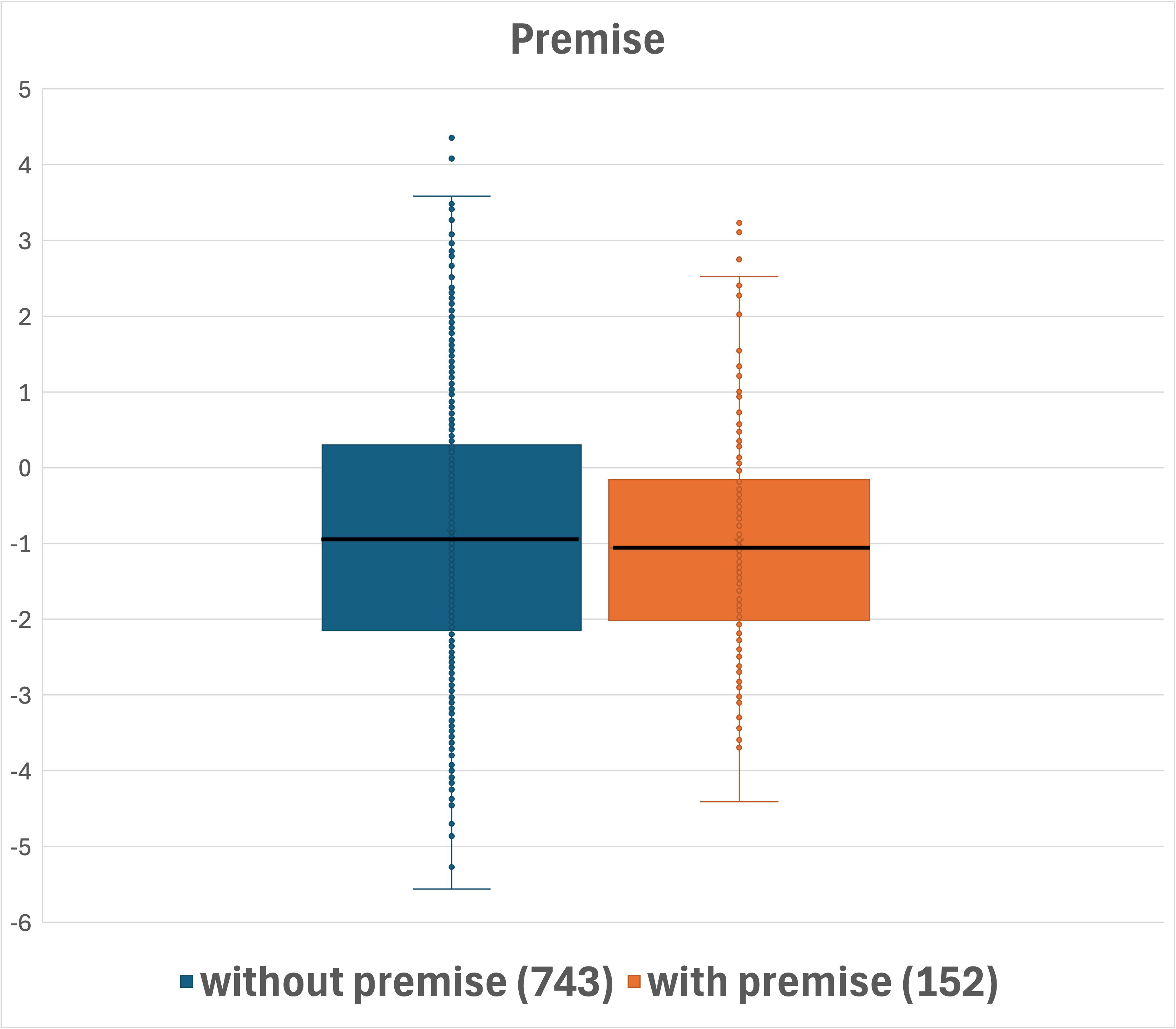}
    \end{subfigure}%
    \\ 
    \begin{subfigure}[b]{0.33\textwidth}
        \includegraphics[width=\textwidth]{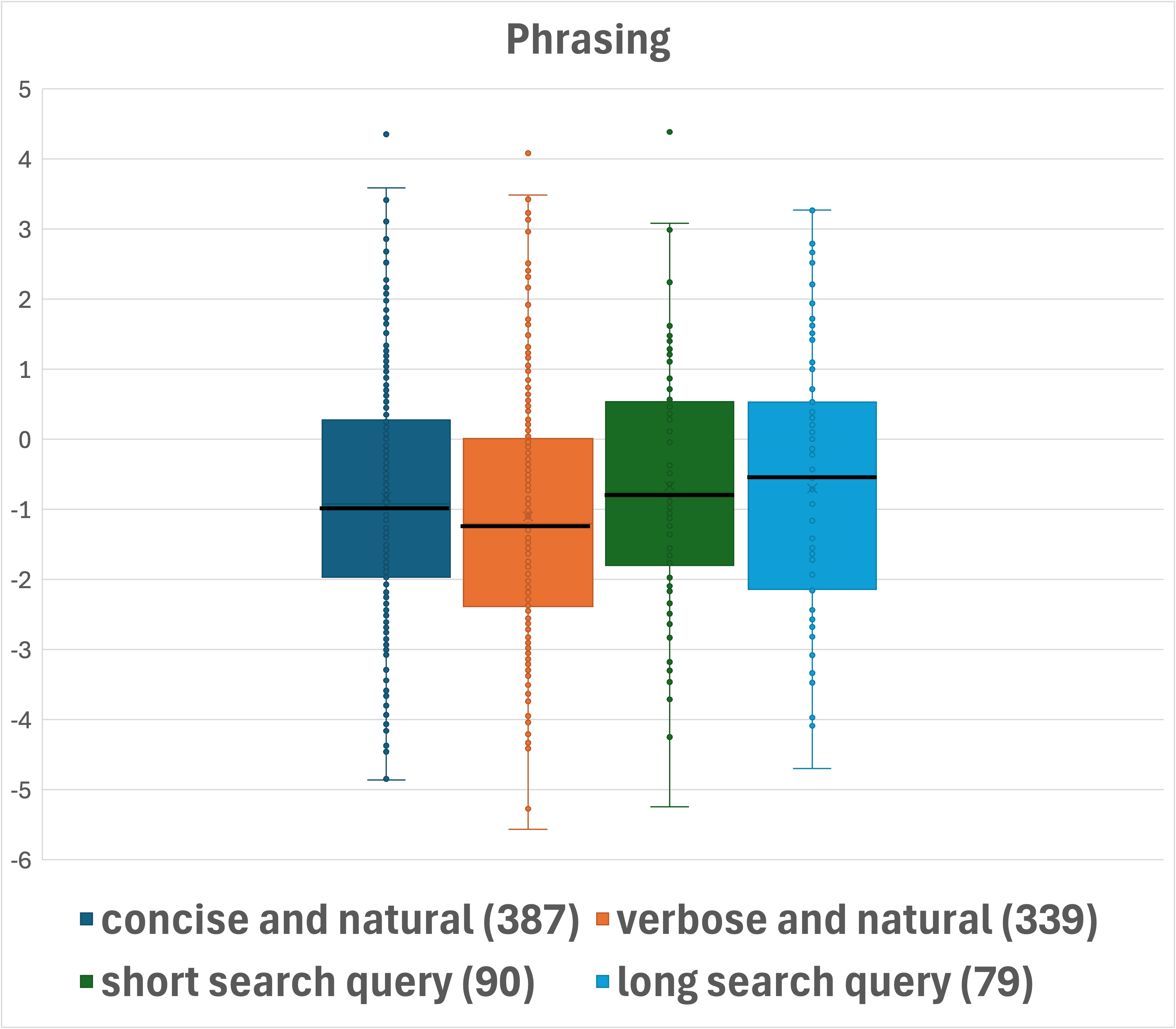}
    \end{subfigure} %
    \begin{subfigure}[b]{0.33\textwidth}
        \includegraphics[width=\textwidth]{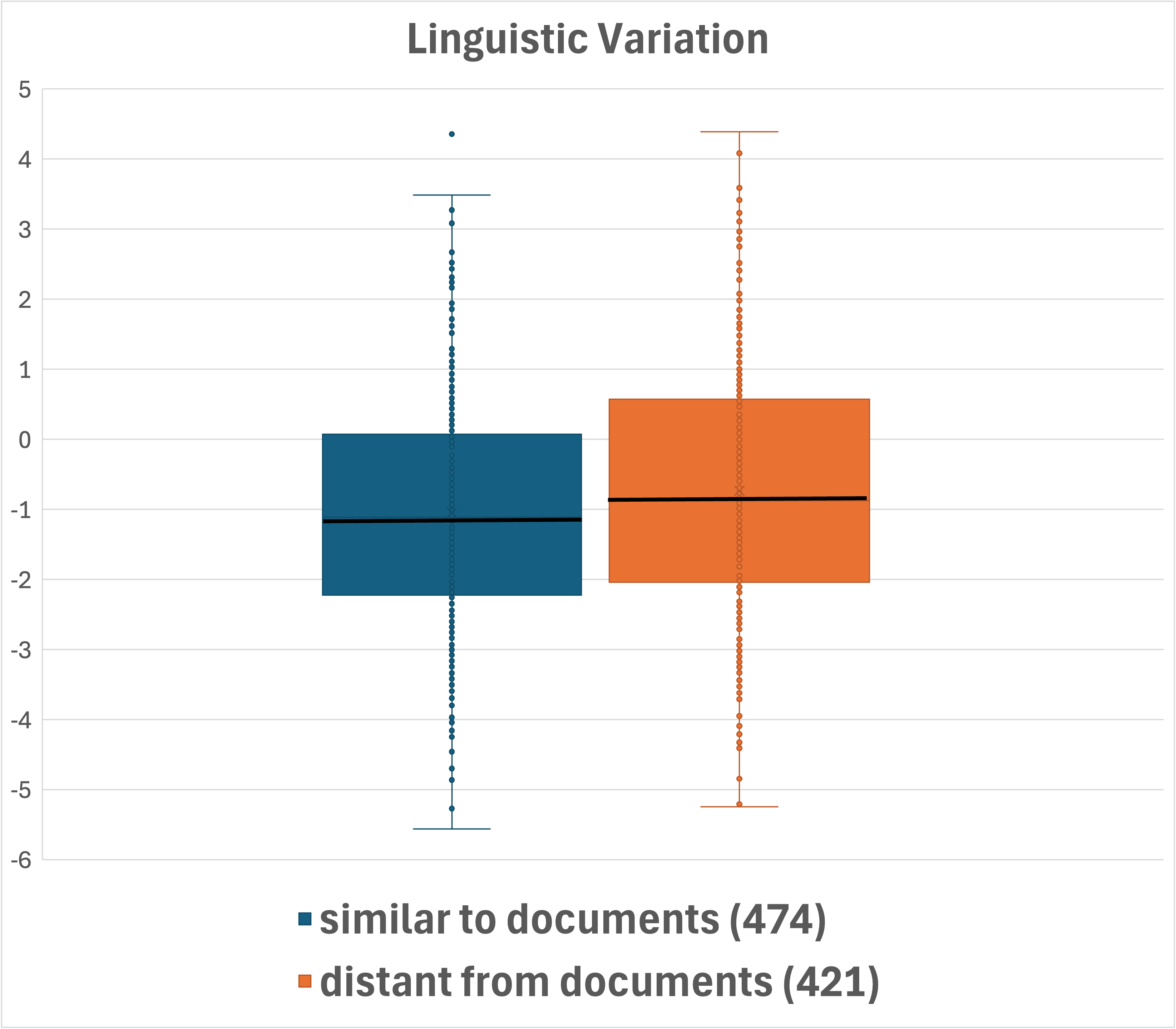}
    \end{subfigure}%
    \begin{subfigure}[b]{0.33\textwidth}
        \includegraphics[width=\textwidth]{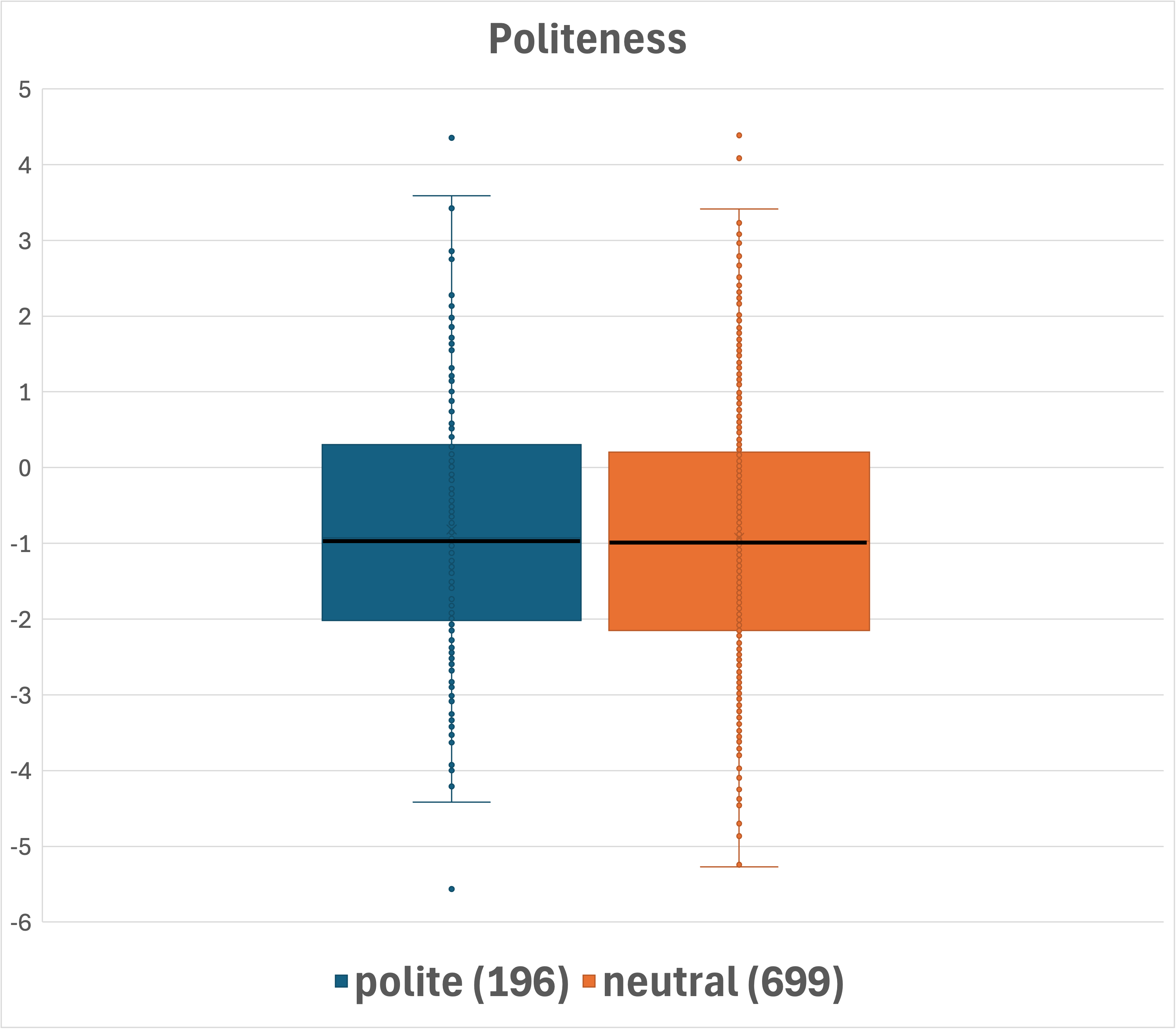}
    \end{subfigure}%
    \\
    \begin{subfigure}[b]{0.33\textwidth}
        \includegraphics[width=\textwidth]{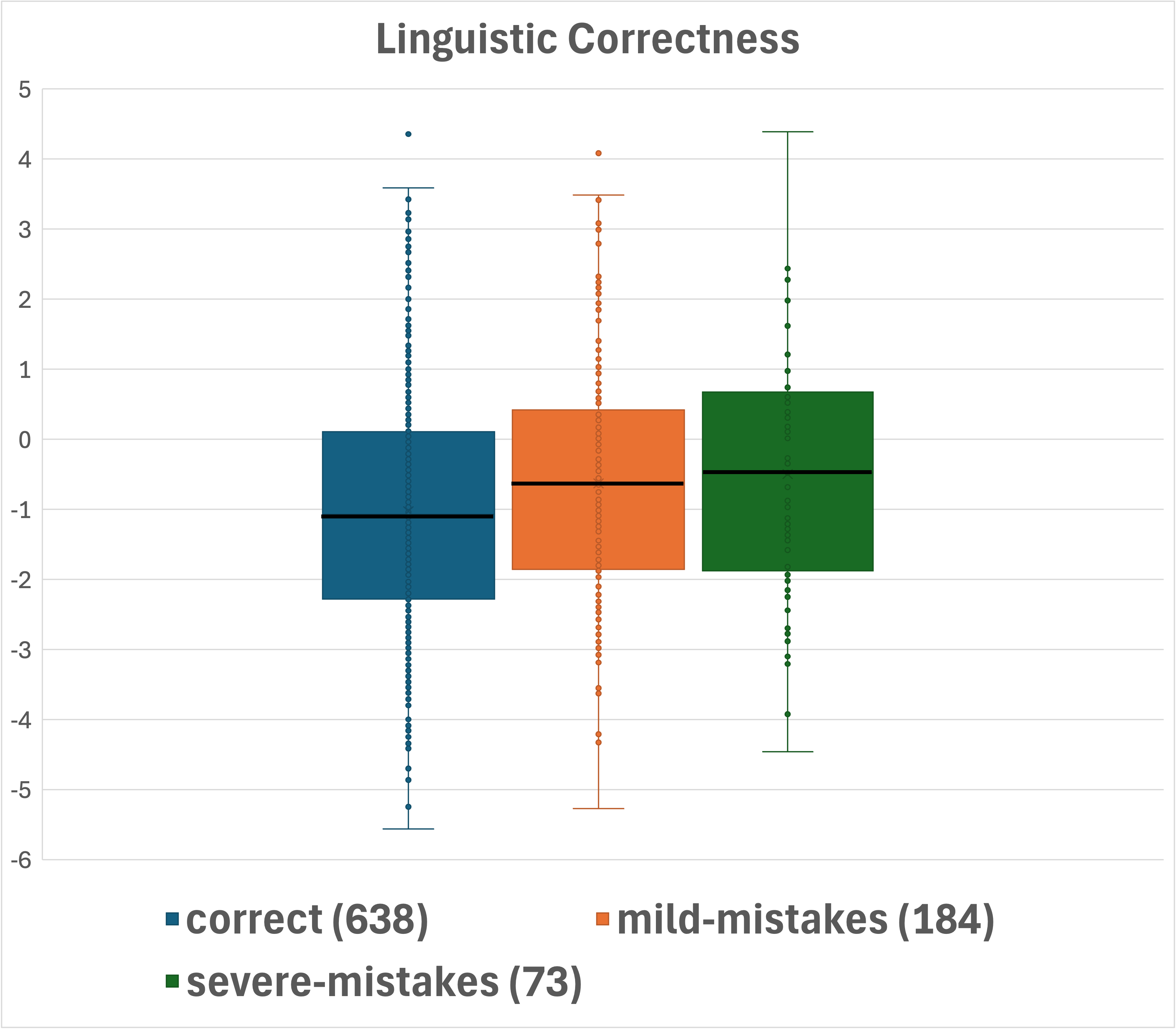}
    \end{subfigure}%
    \begin{subfigure}[b]{0.33\textwidth}
        \includegraphics[width=\textwidth]{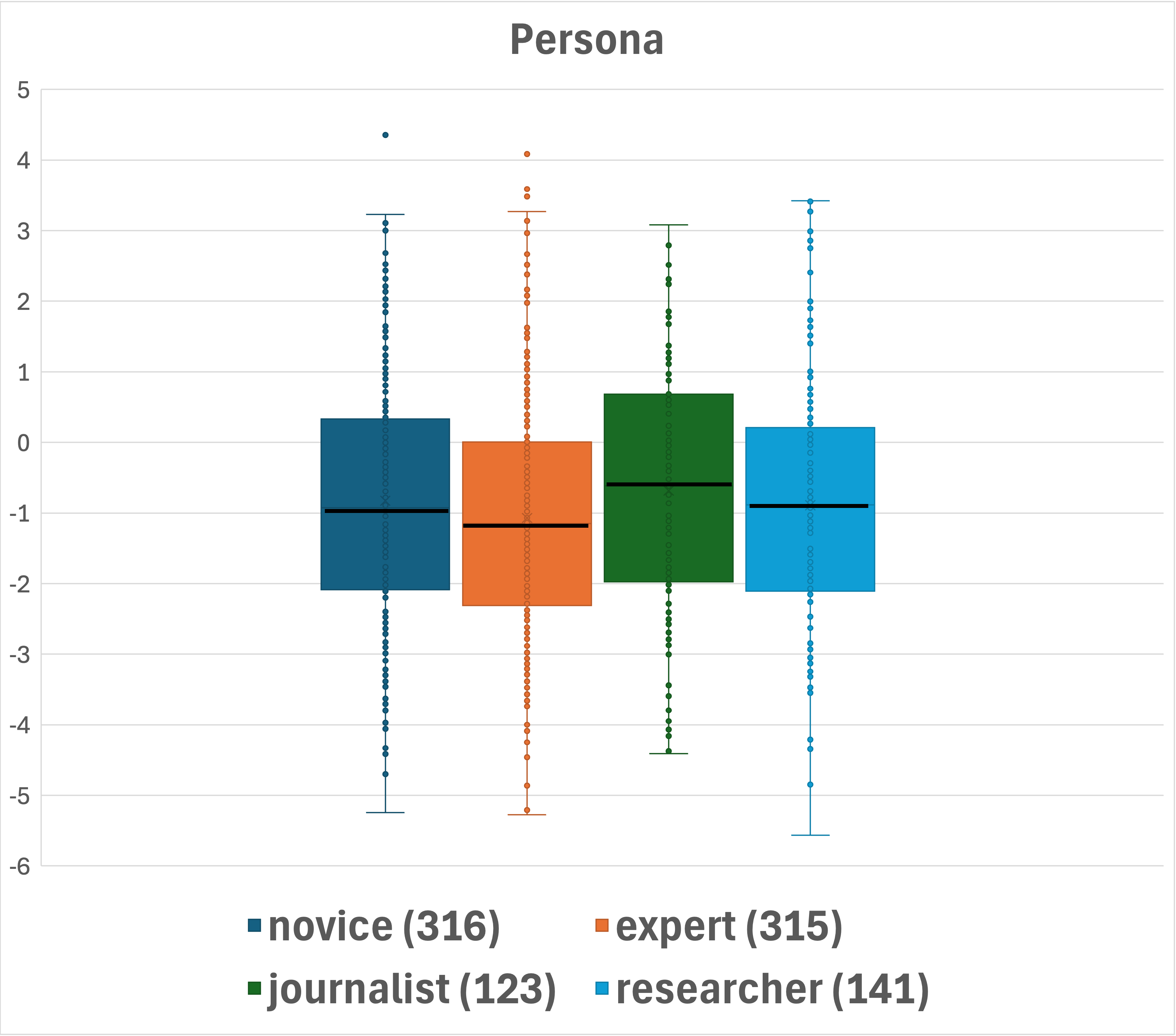}
    \end{subfigure} %
    \caption{Box-plot presentation of the \diff{} distributions across question categorizations. Median values are shown in bold. Number of questions per category is indicated in parentheses.}
    \label{fig:QDiffperCategory}
\end{figure*}

We therefore measure the {\it diff} distribution across the different question categorizations used by DataMorgana for the challenge (see Table \ref{tab:question categories}). Figure \ref{fig:QDiffperCategory}
presents these distributions across the eight categorizations used. The number of questions per category (given in parentheses), is determined by the probability distribution specified within DataMorgana configuration.


Although the differences in average \diff{} across categories are statistically insignificant for most categorizations, the distributional shifts between categories are still observable.
Looking at the Answer Type categorization, \emph{comparison} and \emph{multi-aspect} questions emerge as the most challenging. This outcome is expected, as both types require synthesizing information from multiple documents, unlike other answer types, which rely on a single document.
Interestingly, \emph{Yes/No} questions also appear to be relatively more difficult. We hypothesize that this is due to their binary nature, which often requires implicit reasoning, especially when the correct answer (Yes/No) is not explicitly stated in the source text.

Similarly, in the \emph{Answer Style} categorization, \emph{concise-answer} questions are relatively more difficult for instruction-tuned LLMs, which are typically trained to produce elaborated responses. A similar pattern is observed in the \emph{Phrasing} categorization, where natural questions are easier than search (keyword-based) questions, which are inherently more ambiguous due to their compressed format.

For \emph{Linguistic Variation}, the analysis reveals that questions which are semantically similar to their documents are easier than those that are dissimilar. This is reminiscent of reported work on query difficulty estimation for ad hoc retrieval~\cite{carmel2006what}, where the ``distance'' between the query and the retrieved documents strongly influences its difficulty.  
\emph{Premise} and \emph{Politeness} do not seem to affect difficulty, while for \emph{Linguistic Correctness}, as expected, the severity of typos embedded within the questions increases their difficulty.  Finally, for \emph{User Persona}, \emph{expert} questions are slightly easier than \emph{novice} questions, likely because they contain specific terminology that facilitates the retrieval process.


\subsection{Linguistic Diversity}
\begin{table*}[th]
\centering   

\begin{tabular}{lccccc}
      \multirow{2}{*}{\textbf{Benchmark}}                 &    \textbf{NGD}                        &   \textbf{PoS-CR}                                     & \textbf{embbed.}            & \textbf{length}              & \textbf{question}\\
      &\textbf{($\uparrow$)}  & \textbf{($\downarrow$)}        & \textbf{HS ($\downarrow$)} &\textbf{entropy ($\uparrow$)} & \textbf{length}\\
    \toprule
    TriviaQA \cite{joshi-etal-2017-triviaqa}  & 3.045 & \textbf{5.167} & 0.055 & 3.085 & 14.301 \\
    PopQA \cite{mallen2023not}  & 1.952 & 10.711 & 0.214 & 1.691 & 6.669 \\
    WebQuestions \cite{berant-etal-2013-semantic} & 2.762 & 6.175 & 0.157 & 1.848 & 6.736 \\
    SimpleQA \cite{wei2024measuringshortformfactualitylarge}  & 2.755 & 6.184 & 0.110 & 3.126 & 16.399 \\
    Natural Questions \cite{kwiatkowski-etal-2019-natural} & 2.841 & 5.373 & \textbf{0.017} & 1.518 & 9.029 \\
    \midrule
    LiveRAG & \textbf{3.062} & 5.220 & 0.061 & \textbf{3.207} & 15.297 \\
    \hline
\end{tabular}
    \caption{Linguistic diversity metrics for the \liverag{} benchmark, and several popular QA benchmarks; $\uparrow$/$\downarrow$ mark higher/lower-is-better.}
    \label{tab:diversity}    
\end{table*}
Diversity is a key characteristic of any benchmark designed to evaluate system performance, as it  broadens the spectrum of challenges and scenarios a system may encounter in real-world deployments. A diverse benchmark helps models to generalize across different question types and is more likely to include edge cases and uncommon styles, increasing the challenge for models trained primarily on homogeneous data  \cite{han2014big}.

To evaluate the diversity of the benchmark, we adopt several metrics developed for text generation tasks \cite{shaib2025standardizingmeasurementtextdiversity}, which capture general linguistic aspects, such as lexical, syntactic, and semantic diversity. Table~\ref{tab:diversity} presents the linguistic diversity of the \liverag{} questions and of equal-size samples (i.e., 895 questions) from some popular QA benchmarks. The \liverag{} benchmark achieves the highest lexical diversity, as measured by NGD --- the fraction of distinct n-grams (up to 4) over the total number of n-grams.  Additionally, the \liverag{} questions  also reach the highest length entropy --- the entropy of question length distribution in the benchmark. To compute syntactic diversity, the benchmark questions are first converted into their \textit{Part-of-Speech} (PoS) tag sequences, and then the compression ratio, PoS-CR, is defined as the ratio between the size of the file containing the PoS sequences to the size of its compressed gzip version. According to this metric, only TriviaQA is more diverse than the \liverag{} benchmark\footnote{
TriviaQA contains many questions  with unusual syntactic patterns (e.g., \textit{``A Russian rouble is divided into 100 … .what?''}) that highly contribute to its syntactic diversity. 
}.

To assess semantic diversity, we compute the Homogenization Score (embeddings-HS), which measures the average pairwise cosine similarity between question embeddings\footnote{Question embeddings are obtained using the \textit{MiniLM} sentence encoder {\url{https://huggingface.co/sentence-transformers/all-MiniLM-L6-v2}}}. 
The \liverag{} benchmark achieves competitive diversity results, despite all questions being generated from a limited set of topics (see Section \ref{sec:dm_input}), a factor that would typically reduce semantic diversity.

These results suggest that \liverag{} is generally more diverse than widely used benchmarks, offering a robust framework for evaluating RAG systems. Moreover, the diversity analysis complements and reinforces the difficulty analysis presented in Section \ref{sec:analysis}, as diversity is inherently linked to difficulty since higher language variation often requires diverse reasoning strategies to answer the question.

\section{Limitations}
\label{sec:limitations}

The synthetic Q\&As in the \liverag{}  benchmark are not direct reflections of actual user needs, but rather projections of anticipated future needs, as approximated through the DataMorgana categorizations. Consequently, conclusions about system performance in real-world scenarios derived from this benchmark should be interpreted with caution. Moreover, although the benchmark Q\&As appear natural and well formulated, the corresponding answers are automatically generated based on one or two source documents, while other documents in the corpus may offer more accurate or even contradictory answers.  Such discrepancies can lead to high-quality responses being mistakenly evaluated as incorrect due to divergence from the designated ``ground truth''.

Furthermore, the IRT-based {\it diff} and {\it disc} scores are calculated based on the responses of systems that participated in the \liverag{} Challenge. Since all these systems utilized a RAG-based architecture based on Falcon3 for answer generation, these scores may reflect biases where certain factors contributing to question difficulty or discrimination are specific to such models. 
Despite these limitations, our empirical analysis supports the reliability of these scores as indicators of question difficulty.

\section{Concluding Remarks}
\label{sec:conclusions}

In this paper, we introduced the \liverag{} benchmark -- a publicly available dataset 
based on the dataset used in the SIGIR'2025 \liverag{} challenge. It enriches the Q\&A pairs by including the average and standard deviation of Correctness scores achieved by participating teams for each question, as well as difficulty and discriminability scores derived from IRT analysis, which can serve as proxies for question difficulty and discriminative power, respectively. Our preliminary analysis explored the distribution of question difficulty across various dimensions and demonstrated the reliability of these metrics. We observed that highly difficult questions posed significant challenges to all participating systems, as well as to a range of LLMs of different sizes. While such difficult questions may be less effective at differentiating between systems, they expose important limitations of current RAG approaches and highlight key directions for future research.
\bibliographystyle{unsrt}
\bibliography{biblio}

\appendix

\newtcolorbox{promptbox}[1]{breakable=false,left=2pt,right=2pt,fontupper={\small\ttfamily}, fontlower={\small\ttfamily},title={#1},fonttitle={\small},skin=enhanced jigsaw,colframe=black,colback=gray!5!white,size=fbox}

\section{Prompts}\label{appendix:prompts}
\subsection{Topic generation prompt}\label{appendix:prompts:topic_generation}
\begin{promptbox}{Topic Generation}
Generate a list of <n> high-level topics that can be suitable for a knowledge-based question-answer system. The topics should be diverse and concise.
Return the topics as a python list. Do not output any preamble or explanation.
\end{promptbox}

\begin{promptbox}{Sub-Topic Generation}
Generate <n> sub-topic of this topic: <topic>. Each sub-topic must be suitable for a knowledge-based question-answer system. The sub-topics should be diverse and concise.
Return the topics as a python list. Do not output any preamble or explanation.
\end{promptbox}

\subsection{Document filtering prompt}\label{appendix:prompts:document_filtering}
\begin{promptbox}{Document filtering}
Your task is to give a score on a scale of 1-5 for the following document based on the following aspects:\\
1) factuality: The document contains factual information about an entity/concept that is expected to be found on Wikipedia (or similar Encyclopedias). The document is appropriate for creating an open domain question from.\\
2) interest: The document contains knowledge which might be interesting or useful to someone.\\
3) credibility: The document is credible and can be trusted as reliable, accurate, and authoritative. Examples of documents that are NOT credible include promotional materials, heavily biased political materials, narratives, diaries, resumes, job posting, subjective reviews, essays without supporting evidence, or unanswered questions.\\
4) toxicity: The document contains harmful, offensive, or inappropriate language, such as hate speech, harassment, or extreme negativity.\\
5) sexuality: The document contains sexual content.\\
6) outdated: The document contains outdated content.\\
Return your answer without any preamble in the following JSON format: \{"factuality": <score>, "interest": <score>, "credibility": <score>, "toxicity": <score>, "sexuality": <score>, "outdated": <score>\}\\
Document: <document>
\end{promptbox}

\subsection{Query generation prompt}\label{appendix:prompts:query_generation}
The following prompt is used to generate search queries that retrieve complementary documents to the seed document during the multi-document question generation process.
\begin{promptbox}{Query generation}
Generate <n> questions satisfying the following characteristics:\\
<question category description>\\
It must be possible to answer the questions by using information from two documents, which we will call DOC-A and DOC-B.\\
The required information from DOC-A must not appear in DOC-B and vice versa.
In addition, for each question you must generate a query that can be used in a search engine to retrieve DOC-B.\\
Generate <n> question-query pairs. Write each pair in a new line, in the following JSON format:\\
\{ "question": "<the generated question>", "search query": "<the search query for DOC-B>" \}\\
Do not provide any preamble or explanation.\\
\\
\#\#\#\ DOC-A:\\
<d1>
\end{promptbox}

\subsection{Document selection prompt}\label{appendix:prompts:document_selection}
The following prompt is used to select a complementary document from search results during the multi-document question generation process.
\begin{promptbox}{Document selection}
I need to generate a question having the following characteristics:\\
<question category description>\\
It must be possible to answer the question by using information from two documents, which we will call DOC-A and DOC-B. 
The required information from DOC-A must not appear in DOC-B and viceversa.\\
You will receive DOC-A and a numbered list of candidate documents from which you must select DOC-B.
First provide a short reasoning for your decision, and then your pick.\\
You must respond in the following JSON format: 
'\{"reasoning": <short reasoning>, "doc number": <the number of the selected document>\}'\\
If no document is suitable for generating a question with the expected characteristics, write "null" in "doc number".\\\\
\#\#\#\ Document A:\\
<d1>\\
\\
\#\#\#\ Candidate documents:\\
\#\#\#\ Document 1:\\
<candidate document 1>\\
\#\#\#\ Document 2:\\
<candidate document 2>\\
...\\
\#\#\#\ Document m:\\
<candidate document m>\\
\end{promptbox}

\section{Benchmark Description}
\label{app:description}

\begin{table*}[t]
    \begin{tabular}{l|l}
{\bf field-name} & {\bf field-value} \\
\toprule
Question: & How deep can fish survive in the ocean trenches? \\
Answer: & \textcolor{red}{Fish can survive up to 8,100 meters deep}. \textcolor{blue}{Beyond this depth, no fish live},\\               
        & \textcolor{blue}{possibly due to their inability to cope with the highest pressures.} \\
Supporting Docs: & <urn:uuid:a102a6cb-a608-493c-928f-d32a0da4dbf6> \\ 
DM Categories: & unspecified,  factoid, concise and natural, correct, similar to documents, \\ 
               & polite,  without premise, novice \\
ACS: & 0.876 \\
{\em diff}:  & -1.33 \\
{\em disc}: & -0.29 \\
\midrule
Question: & Based on temperature considerations alone, is March considered a \\
          & suitable month to perform the final pruning of grape vines? \\
Answer:   &  \textcolor{red}{Yes.The first week of March is a good time for the final pruning of grapes}, as this is\\
          & when \textcolor{blue}{you can prune the remainder of last year's growth down to about one or 2}  \\
          &  \textcolor{blue}{inches long for spur production, or 12 to 15 inches long for cane production}. \\
Supporting-docs: & <urn:uuid:b5d19fcb-1711-4f9f-82cf-f81403382444> \\
DM Categories: &  concise-answer, yes/no, verbose and natural, correct, distant from documents, \\
               & neutral, without premise, novice \\
ACS: & 0.089 \\
{\em diff} & 1.234 \\
{\em disc}: & -0.162 \\
\midrule
Question: & Would you please explain how both Amália Rodrigues and Gene Chandler demonstrated \\
   & good artistic intuition in choosing their musical direction? \\
Answer: & Both artists showed strong artistic intuition in their career choices.  \textcolor{red}{Amália Rodrigues} \\
   &  \textcolor{red}{had a strong intuition and good taste in selecting her repertoire, consistently }\\
   &  \textcolor{red}{choosing compositions from great Portuguese composers} and  \textcolor{blue}{later expanding into} \\ 
   &  \textcolor{blue}{classical poetry with Oulman}.  \textcolor{red}{Gene Chandler} demonstrated his intuition when he  \textcolor{red}{promoted} \\ 
   &  \textcolor{red}{the flip side 'Rainbow' against his label's wishes to push 'You Threw a Lucky Punch,'} \\
   & a decision that proved smart as  \textcolor{blue}{'Rainbow' became more successful and helped establish him }\\
    &  \textcolor{blue}{as a serious artist beyond his 'Duke of Earl' hit}. \\
Supporting Docs: & <urn:uuid:8269372f-be5c-4212-8830-c7e0ca1b295e>, \\
                 & <urn:uuid:af9d46db-12a9-41d8-9687-1b4232404418> \\
DM Categories: & unspecified, comparison, verbose and natural, correct, similar to documents \\
               & polite, without premise, expert\\
ACS: &0.277 \\
{\em diff} & 1.317 \\
{\em disc}: & -0.184 \\
\bottomrule
\end{tabular}
    \caption{\label{ta:examples} A few examples from the benchmark: Top: an easy question. Middle: a difficult question. Bottom: a highly difficult (multi-doc ) question. In red - answer direct claims. In blue - answer useful claims.}
\end{table*}

In this appendix, we describe the \liverag{} Benchmark, hosted on the open Hugging Face platform\footnote{
\label{footnote: liverag benchmark}\url{https://huggingface.co/datasets/LiveRAG/Benchmark}
}, which includes 895 Q\&As. 
Table \ref{ta:examples} presents a few illustrative examples from the benchmark.
Each entry in the benchmark includes the following fields:
\begin{itemize}[leftmargin=9pt]
    \item \emph{Question:} the question generated by DataMorgana.    
    \item \emph{Answer:} the corresponding answer generated by DataMorgana.
    Since the answer is generated using a strong LLM, based on selected documents from the external corpus, it is treated as the ``ground truth'' answer for the given question. However, this may lead to inconsistencies when other documents in the corpus provide alternative valid answers. In such a case, a correctly generated answer might be incorrectly judged as wrong. To minimize these issues, we manually filtered out problematic items from the benchmark. Nevertheless, it is important to note that the provided answer is not necessarily the only valid answer to the question.
    \item \emph{Supporting documents:} the list of Fineweb-10BT documents used for Q\&A generation. The list contains either a single document (for single-document questions) or two documents (for multi-document questions). Each document includes its FineWeb-10BT document ID and its full content.
    \item \emph{Answer claims:} The \liverag{} official evaluation \cite{carmel2025sigir2025liverag} is based on comparing the generated answer's claims to the vital claims present in the reference answer. For that, we extract all claims from the answer and classify them into three categories: 1) \textit{Direct} -- the claim directly corresponds to answering the question; 2) \textit{Useful} -- the claim is useful for answering the question; and 3) \textit{Useless} -- the claim is unrelated or unhelpful for answering the question. The list of answer claims with their corresponding classifications is provided to support the evaluation process.
    \item \emph{Session:} Indicates the Live Challenge Day session in which the question appeared (``First'' - first session, ``Second'' - second session, and "Both" - both sessions, i.e., a shared question). We provide this information mostly to help benchmark users compare their scores against those achieved by competitors in each particular session. 
    \item \emph{DataMorgana configuration:} The eight question categories used by DataMorgana for question generation.
    \item \emph{Average Correctness Score (\acs{}):} The average Correctness score
    across all \liverag{} systems which answered the question. 
    \item \emph{Standard deviation of \acs{} (\acs{}\_std):} The standard deviation of the Correctness scores of all \liverag{} systems which answered the question.
    \item \emph {IRT parameters:} the (\diff{}, \disc{}) parameters learned by the IRT-2PL model (see Section \ref{sec:irt-model}).
\end{itemize}



\end{document}